\documentclass[10pt,twocolumn,letterpaper]{article}
\usepackage[pagenumbers]{cvpr}
\usepackage{amsfonts}
\usepackage{mathtools}
\usepackage{multirow}
\usepackage{enumitem}
\usepackage{tikz}
\usetikzlibrary{arrows.meta,positioning,fit,calc}
\usepackage{tabularx}
\usepackage{array}
\usepackage{placeins}
\usepackage{needspace}
\definecolor{cvprblue}{rgb}{0.21,0.49,0.74}
\usepackage[pagebackref,breaklinks,colorlinks,allcolors=cvprblue]{hyperref}
\def\confName{CVPR}
\def\confYear{2027}
\newcommand{\method}{ClassReST}

\title{Beyond Domain Alignment: Restoring Class Discrimination in Open-Set Test-Time Adaptation}

\author{%
Yunkai Bian\textsuperscript{1} \quad
Zhun Luo\textsuperscript{2} \quad
Song Tan\textsuperscript{2} \quad
Congfang Hu\textsuperscript{3}\\
Lixin Lin\textsuperscript{1} \quad
Hao Zhang\textsuperscript{1,*} \quad
Wei Luo\textsuperscript{4,*}\\[0.6em]
\textsuperscript{1}Central South University, Changsha, China\\
\textsuperscript{2}Mango TV, Changsha, China\\
\textsuperscript{3}Xiangtan University, Xiangtan, China\\
\textsuperscript{4}Deakin University, Melbourne, Australia\\[0.3em]
{\tt\small
254512044@csu.edu.cn \quad hao@csu.edu.cn \quad wei.luo@deakin.edu.au}\\
}

\begin{document}
\maketitle
\begin{abstract}
When a visual classifier is deployed, environmental changes can corrupt its inputs and cause familiar objects to be misclassified. Open-set continual test-time adaptation (OCTTA) seeks to maintain recognition of known classes while rejecting unknown inputs as visual conditions change. Existing approaches compensate for corruption by aligning feature distributions or preserving relationships among examples. These constraints leave class correspondence unresolved: our counterexample preserves feature statistics and similarities while known-class assignments remain incorrect. Diagnostics of native DOCO show that compensation updates can reduce the adaptation objective and worsen individual correct-class margins.

We propose ClassReST (Class-Discriminative Restoration of Spatial Tokens), which learns to correct these residual class confusions using labelled clean–corrupted source image pairs. By comparing the classifier's responses to clean and corrupted views, we train a spatial restorer to recover the class relationships observed in clean images. The learned corrections depend on the input and operate on intermediate visual features, allowing the restorer to address different corruptions without knowing their types. During deployment, the restorer remains frozen and corrects predictions under evolving, history-stabilised domain-compensation states; its outputs do not enter the online adaptation objective. It requires neither target labels nor clean test references. Controlled experiments show that the learned spatial corrections recover residual class errors under matched online states, while the complete system improves known-class recognition and unknown detection across changing corruption conditions.

\end{abstract}

\section{Introduction}
\label{sec:intro}

When a visual classification model is deployed, the conditions under which it operates often differ from those encountered during training. Changes in lighting, weather, or image quality can introduce noise, blur, and other corruptions that make familiar objects harder to recognise~\cite{hendrycks2019corruptions}. The model may also encounter objects outside its training categories~\cite{liu2020energy}. Open-set continual test-time adaptation (OCTTA) addresses these challenges by adapting the model to changing visual conditions using unlabelled inputs, while maintaining its ability to recognise known-class (in-distribution, ID) inputs and reject unknown-class (out-of-distribution, OOD) inputs~\cite{yang2026doco}.

\begin{figure*}[t]
    \centering
    \includegraphics[width=\textwidth]{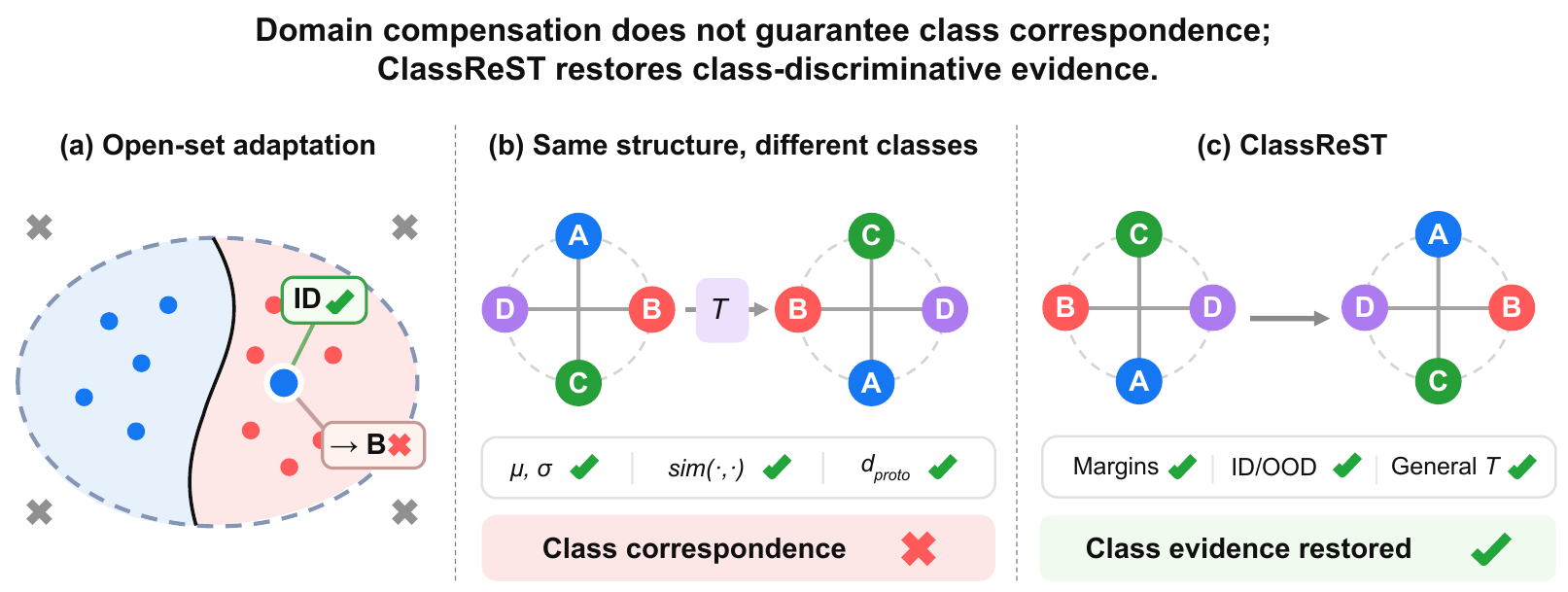}
%    \caption{
%        \textbf{Motivation: Beyond domain compensation.}
%        (a) Standard OCTTA processes sequential corruptions, whereas SDC-OCTTA introduces dynamic class--corruption associations within mixed batches. (b) DOCO compensates domain shifts but retains class-discriminative errors. (c) \method{} restores class evidence through source-trained spatial correction. On SDC-OCTTA, \method{} improves the H-score over DOCO by 6.2 points. Domain and decision geometries are schematic; the methods use distinct online states. }
    
%    \caption{
%        \textbf{Motivation: Domain compensation does not guarantee class recovery.}
%        (a) Corruption shifts target features away from source statistics.(b) DOCO reduces domain discrepancy but retains ID classification errors.(c) \method{} restores class-discriminative evidence through source-trained spatial correction.The geometries are schematic, and the methods use distinct online states.}

    \caption{An input can be routed as ID yet assigned to the wrong known class. The schematics illustrate that feature statistics ($\mu,\sigma$), pairwise similarity and nearest-prototype distance ($d_{\mathrm{proto}}$) can be preserved while class correspondence changes. The restoration panel depicts the learning goal: use clean-referenced class relationships to guide a spatial correction towards the correct class.}

    \label{fig:motivation}
\end{figure*}

A common approach is to compensate for the changes that corruption introduces into the model's feature representations. For example, DOCO~\cite{yang2026doco} encourages features from incoming images to match source-domain statistics while preserving relationships among the features. These constraints describe feature statistics and geometry without determining correspondence to fixed source-class labels. An image may still be recognised as belonging to a known category but assigned to the wrong class (\cref{fig:motivation}).

We establish this limitation of the compensation constraints through a mathematical counterexample. Diagnostics of native DOCO further show that reducing its adaptation objective need not improve correct-class margins for individual known-class samples. We therefore investigate guidance available before deployment: labelled clean--corrupted source pairs reveal how corruption changes the relative scores of the correct class and its competitors.

We propose \method{}, which learns input-dependent spatial corrections using only these paired source observations. A small network adjusts intermediate visual features towards the class relationships observed in clean images. Once trained, the network remains frozen and corrects predictions under evolving, history-stabilised domain-compensation states, without test labels, clean test counterparts or knowledge of the corruption type.

%To address it, we develop \method{}, which learns to correct corrupted visual features using labelled clean--corrupted source pairs. 
Matched-state experiments isolate residual class-error correction by the spatial restorer, while complete-system evaluations show improvements in known-class recognition and unknown detection under conventional and changing mixed-corruption conditions.

\section{Related Work}
\label{sec:related}

\paragraph{Continual and open-set adaptation.}
Continual test-time adaptation uses unlabelled target predictions to limit error accumulation and forgetting~\cite{wang2022cotta,niu2022eata,niu2023sar}.
Open-set adaptation separates known and unknown inputs to guide updates~\cite{gao2024unient,yang2026doco}.
DOCO compensates distribution shifts using this separation, cached source statistics and constraints on target feature geometry~\cite{yang2026doco}.

\paragraph{Preserving and recovering class discrimination.}
TCA constrains relationships among pseudo-labelled class centroids~\cite{ni2025tca}, while AEA guides energy alignment using virtual logits from classifier weights for predicted classes~\cite{choi2025aea}.
Corruption can make their inferred class assignments unreliable.
DDA projects corrupted inputs towards the source domain using a source-trained generative model~\cite{gao2023dda}.
CleanRelation matches clean-reference margins, relating it to response-based distillation~\cite{hinton2015distilling} and clean-to-noisy teacher--student learning~\cite{hong2021student}.
\method{} uses labelled clean--corrupted source pairs to learn intermediate spatial corrections for residual class confusion after domain compensation.
The input-dependent restorer remains frozen under evolving compensation states without clean target references or target labels.

\newtheorem{proposition}{Proposition}
\section{Problem Formulation and Analysis}
\label{sec:theory}

Let $f_\theta=h_\theta\circ\phi_\theta$ be the source-trained classifier, with feature extractor $\phi_\theta$, readout $h_\theta$ and known-class label space $\mathcal Y^S$. Pretrained parameters $\theta$ stay fixed; prompt parameter vector $p$ adapts. For image $x$, $\ell(x;p)$ is the logit vector and $\ell_c(x;p)$ its class-$c$ component; $y\in\mathcal Y^S$ is the true known-class label and $\hat y=\arg\max_{c\in\mathcal Y^S}\ell_c(x;p)$ its prediction. The unlabelled OCTTA stream crosses successive target domains $T_i$ without state resets, with input distribution~\cite{yang2026doco}
\begin{equation}
P_{\mathrm{test}}^{T_i}
=(1-\kappa)P_{\mathrm{ID-C}}^{T_i}
+\kappa P_{\mathrm{OOD-C}}^{T_i},
\label{eq:octta-mixture}
\end{equation}
Here $P_{\mathrm{ID-C}}^{T_i}$ and $P_{\mathrm{OOD-C}}^{T_i}$ are distributions of corrupted known- and unknown-class inputs, and $\kappa\in[0,1]$ is the unknown proportion. Adaptation sees neither class labels nor ID/OOD indicators; we evaluate known-class accuracy and unknown detection.

\subsection{What domain compensation cannot determine}
\label{sec:theory-domain}
We first ask whether source statistics and pairwise feature geometry determine class correspondence. On the current inferred ID subset, DOCO~\cite{yang2026doco} minimises the domain-compensation objective $\mathcal L_D$ over $p$:
\begin{equation}
\begin{aligned}
\mathcal L_D(p)&=\mathcal L_{\mathrm{stat}}(p)
                   +\beta\mathcal L_{\mathrm{reg}}(p),\\[-1pt]
\mathcal L_{\mathrm{stat}}(p)
 &=\|\hat\mu_p-\mu_S\|_2+\|\hat\sigma_p-\sigma_S\|_2,\\[-1pt]
\mathcal L_{\mathrm{reg}}(p)
 &=\|\operatorname{sim}(Z_p^{\mathrm{ID}})
           -\operatorname{sim}(Z_{\mathrm{raw}}^{\mathrm{ID}})\|_F .
\end{aligned}
\label{eq:doco}
\end{equation}
Here $\mathcal L_{\mathrm{stat}}$ aligns statistics and $\mathcal L_{\mathrm{reg}}$ regularises structure, with regularisation weight $\beta\ge0$. The rows of $Z_p^{\mathrm{ID}}$ and $Z_{\mathrm{raw}}^{\mathrm{ID}}$ are prompted and unprompted final features of the same selected examples. Their prompted coordinate-wise sample mean and standard deviation are $\hat\mu_p,\hat\sigma_p$, with fixed cached source counterparts $\mu_S,\sigma_S$. The operator $\mathrm{sim}$ returns the pairwise row cosine-similarity matrix; $\|\cdot\|_F$ is the Frobenius norm. These constraints preserve pooled statistics and relationships between examples, but need not preserve their correspondence with fixed class labels.

\begin{proposition}[Insufficiency of compensation constraints]
\label{prop:compensation}
Even for known-class examples, vanishing compensation losses do not guarantee correct predictions:
\[
\mathcal L_{\mathrm{stat}}=0,\quad
\mathcal L_{\mathrm{reg}}=0
\quad\not\Rightarrow\quad \hat y=y.
\]
\end{proposition}
\noindent\emph{Proof sketch.}
For a linear classifier with class weight vectors $w_c$ and scalar biases $b_c$, assign the four classes source features $z_y=w_y\in\{e_1,-e_1,e_2,-e_2\}\subset\mathbb R^2$, with standard basis vectors $e_1,e_2$ and $b_c=0$. Set both raw and prompted corrupted representations to $-z_y$. Negation permutes the source multiset, preserving its moments; the two similarity matrices also coincide. Both losses vanish, yet the opposite-class prototype wins for every example. $\square$

The supplement gives the full proof.

\subsection{Why updates need not recover class margins}
\label{sec:theory-mismatch}
The counterexample leaves an optimisation question: do compensation updates improve individual correct-class margins?

At update $t$, fix a diagnostic ID image $x$, its batch and inferred ID subset; $y$ is used only for analysis. Define $m_{y,j}(p)=\ell_y(x;p)-\ell_j(x;p)$ for $j\in\mathcal Y^S\setminus\{y\}$. Let $p_t,p_{t+1}$ be native DOCO's prompt parameters at matched pre-/post-update readouts and $\Delta p_t=p_{t+1}-p_t$. With non-prompt parameters, forward mode and stochastic draws fixed, denote the fixed-batch objective by $\mathcal L_D^{(t)}(p)$. A locally Lipschitz margin gradient gives, for small $\Delta p_t$,
\begin{equation}
\begin{aligned}
\Delta m_{y,j}
&=m_{y,j}(p_t+\Delta p_t)-m_{y,j}(p_t)\\
&=\bigl\langle\nabla_p m_{y,j}(p_t),\Delta p_t\bigr\rangle
+O(\|\Delta p_t\|_2^2).
\end{aligned}
\label{eq:update-margin}
\end{equation}
Loss reduction alone does not require the prompt displacement to align with the margin gradient. For idealised gradient descent at a differentiable point with step size $\eta>0$, $\Delta p_t=-\eta\nabla_p\mathcal L_D^{(t)}(p_t)$; Eq.~\eqref{eq:update-margin} uses the actual prompt displacement under the fixed readout conditions above.

\subsection{What clean--corrupted source pairs reveal}
\label{sec:theory-clean-pairs}
When a corrupted prediction is unreliable, what reference can we obtain before deployment? For a labelled clean source image $(x_s,y)$ and corrupted counterpart $x_c$, let $\ell^s,\ell^c,\ell^r$ denote clean, corrupted and feature-restored corrupted logit vectors under the same frozen source-time compensation state. Their margins define the scalar clean-reference correction target $\Delta^*_{y,j}$:
\begin{equation}
\begin{aligned}
m^s_{y,j}&=\ell^s_y-\ell^s_j,\qquad
m^c_{y,j}=\ell^c_y-\ell^c_j,\\
\Delta^*_{y,j}&=m^s_{y,j}-m^c_{y,j}.
\end{aligned}
\label{eq:clean-reference-change}
\end{equation}
Writing the restored margin as $m^r_{y,j}=\ell^r_y-\ell^r_j$, we can train the correction towards
\begin{equation}
\begin{aligned}
m^r_{y,j}-m^c_{y,j}&\approx\Delta^*_{y,j}\\
&\Longleftrightarrow\quad m^r_{y,j}\approx m^s_{y,j}.
\end{aligned}
\label{eq:clean-reference-target}
\end{equation}
Cross-entropy encourages correct classification across all known classes through $\log(1+\sum_{j\ne y}e^{-m^r_{y,j}})$. Multiple configurations of class margins can achieve this objective. The clean counterpart provides a more specific reference: the relative scores of the true class and its competitors before corruption. CleanRelation uses these margins to guide spatial restoration towards the classifier's clean-image responses (\cref{sec:source_loss}).

\subsection{Why correction direction matters}
\label{sec:class_margin}
Recovering these relationships requires a suitable feature displacement. For a final feature vector $z$ and same-dimensional additive displacement $\delta$, the linear head $\ell_c(z)=w_c^\top z+b_c$ gives
\begin{equation}
 m_{y,j}(z+\delta)-m_{y,j}(z)
 =(w_y-w_j)^\top\delta.
\label{eq:marginlinear}
\end{equation}
For a single image's spatial patch-token matrix $P$ (one row per token), $m_{y,j}(P)$ is the margin through the remaining frozen network and readout, with online state and non-patch tokens fixed at the correction interface. For a small same-shaped displacement $\delta_P$, the first-order response is
\begin{equation}
 m_{y,j}(P+\delta_P)-m_{y,j}(P)
 \simeq\langle\nabla_Pm_{y,j}(P),\delta_P\rangle_F,
\label{eq:direction-local}
\end{equation}
where the matrix inner product sums entrywise products over all tokens and channels.
The margin response depends on displacement direction as well as magnitude. \Cref{sec:method} specifies the learnt correction.

\section{\method{}: Class-Discriminative Spatial Restoration}
\label{sec:method}

\begin{figure*}[t]
    \centering
    \includegraphics[width=\textwidth]{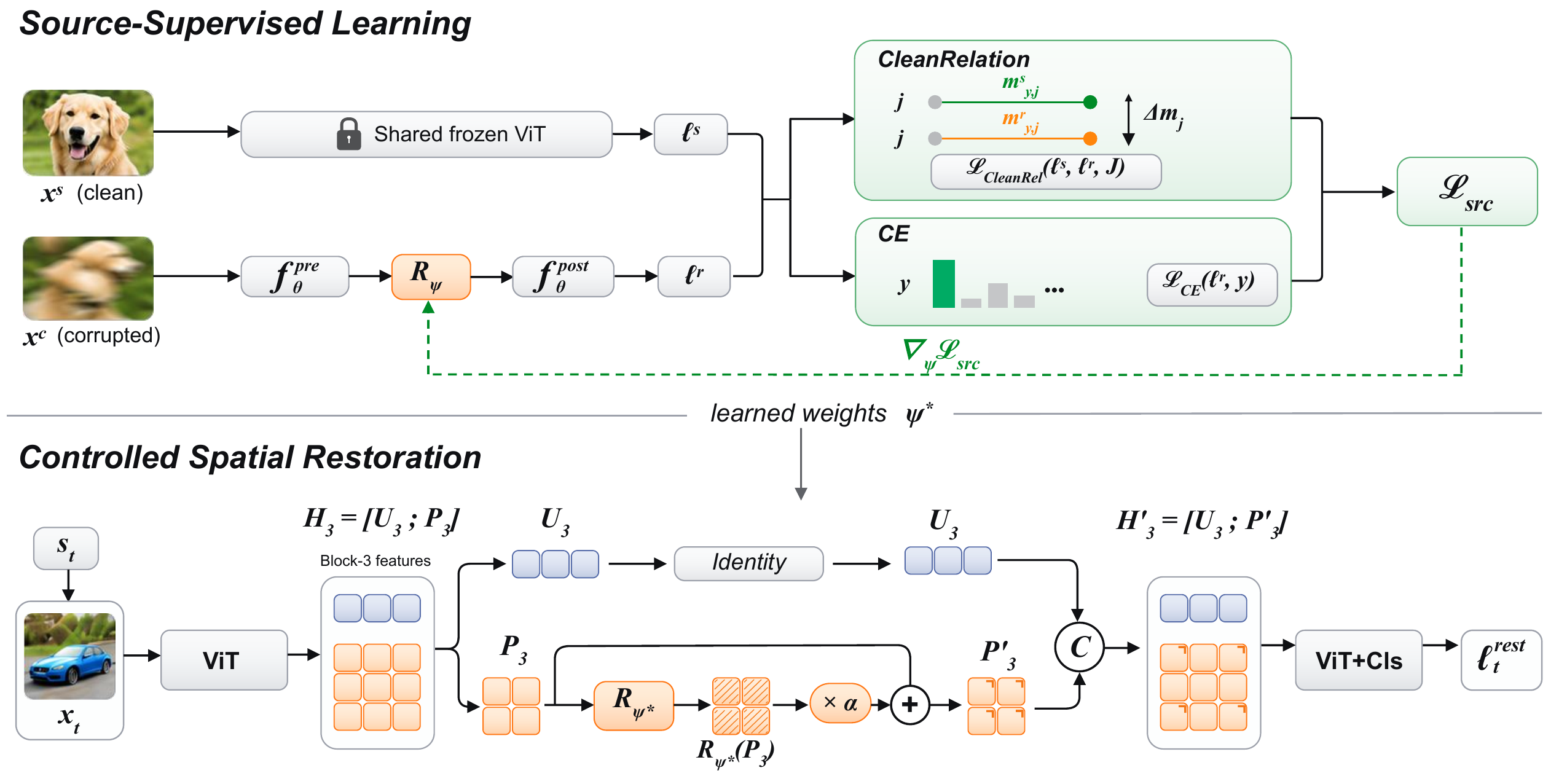}
    \caption{\textbf{Learning and applying class-discriminative spatial correction.}
    Top: clean--corrupted source comparisons provide class-margin references for learning a spatial correction through the frozen classifier. Cross-entropy and CleanRelation update only the restorer. Bottom: the frozen restorer corrects intermediate patch features with source-calibrated strength, and the original suffix and classifier produce predictions under the evolving adaptation state. The evaluated compensation trajectory uses history stabilisation; restoration outputs do not enter its online objective.}
    \label{fig:method}
\end{figure*}

We train \method{} to correct corrupted visual features using class relationships observed in their clean counterparts. A frozen classifier provides clean-reference margins for labelled clean--corrupted source pairs, and a small network learns spatial corrections through that classifier. At deployment, the frozen restorer corrects features under history-stabilised compensation (\cref{fig:method}).

\subsection{Input-dependent spatial correction}
\label{sec:restoration_control}
We correct intermediate spatial features so that the remaining transformer blocks can recompute the visual evidence used for classification. The restorer acts on patch tokens after the third Vision Transformer (ViT) block. Under the current compensation state, write these features as $H_3=[U_3;P_3]$, where $U_3$ contains classification and prompt tokens and $P_3$ contains spatial patches. A network $R_\psi$, with learned parameters $\psi$, predicts a residual of the same shape:
\begin{equation}
P'_3=P_3+\alpha R_\psi(P_3),\qquad
H'_3=[U_3;P'_3].
\label{eq:operator}
\end{equation}
The input features determine the residual's direction and spatial structure, while $\alpha$ controls its application strength separately. These properties shape how the correction changes class evidence through the downstream network (\cref{sec:class_margin}).

Prefix tokens remain unchanged at the correction interface. The original transformer suffix $F_{4:L}$, final normalisation and readout $h_\theta$ produce the restored logits:
\begin{equation}
\ell^r=h_\theta\!\left(\operatorname{Norm}\!\left(
F_{4:L}(H'_3)\right)\right).
\label{eq:readout}
\end{equation}
Here $L$ indexes the final transformer block, and $h_\theta$ includes the original token aggregation and classification head. The backbone and head weights stay fixed. Architectural specifications are given in the supplement.

\subsection{Learning from clean--corrupted source pairs}
\label{sec:source_loss}
For a labelled source image $(x_{s,i},y_i)$ and corrupted counterpart $x_{c,i}$, evaluating both views under the same frozen source-time compensation state gives clean logits $\ell_i^s$ and corrupted logits $\ell_i^c$. Applying the restorer to the corrupted features gives $\ell_i^r$ through \cref{eq:readout}.

We focus on selected high-scoring classes that compete with $y_i$ in either view. Clean predictions identify existing competitors, while corrupted predictions reveal classes that become competitive after degradation. We concatenate the two selected rival lists into $\mathcal J_i$, retaining repeated entries so that a class selected in both views contributes twice. Using the margins from \cref{sec:theory-clean-pairs}, with $m^v_{i,j}=m^v_{y_i,j}$ for source sample $i$, we encourage restored margins to approach their clean counterparts. For a minibatch of $B$ source pairs,
\begin{equation}
\begin{aligned}
\mathcal L_{\mathrm{CleanRel}}
&=\frac{1}{B}\sum_{i=1}^{B}\frac{1}{|\mathcal J_i|}
  \sum_{j\in\mathcal J_i}\\[-3pt]
&\quad\operatorname{SmoothL1}\!\left(m_{i,j}^{r}-m_{i,j}^{s}\right).
\end{aligned}
\label{eq:relation_loss}
\end{equation}
Here $\operatorname{SmoothL1}$ penalises deviation from the clean margin. We combine this term with full-class cross-entropy, using weight $\lambda_{\mathrm{rel}}$:
\begin{equation}
\mathcal L_{\mathrm{src}}
=\mathcal L_{\mathrm{CE}}+
\lambda_{\mathrm{rel}}\mathcal L_{\mathrm{CleanRel}},
\label{eq:lossgroup}
\end{equation}
where $\mathcal L_{\mathrm{CE}}=-B^{-1}\sum_i\log\operatorname{softmax}(\ell_i^r)_{y_i}$. Cross-entropy encourages correct classification across all known classes; CleanRelation complements it by matching selected margins to values supplied by the clean view.

We train one shared restorer with these complementary objectives across simulated source-time compensation states. This exposes the restorer to feature variation from changing compensation states, as encountered across deployment batches. Only $\psi$ is updated, with gradients passing through the frozen suffix and classifier. Source fitting uses the full residual ($\alpha=1$). After training, we select one deployment strength by maximising top-1 accuracy on held-out, image-disjoint source validation data and keep it fixed throughout adaptation. No target labels or target-based calibration are used. The exact rival-selection rule, source-state construction and calibration details are specified in the supplement.

\subsection{Restoration during online adaptation}
\label{sec:method-online}
We evaluate spatial restoration under history-stabilised domain compensation. The frozen restorer corrects intermediate features for individual predictions; history stabilisation moderates compensation updates across successive batches. We isolate restoration's effect through matched-state comparisons with identical states and readout timing.

For an unlabelled batch $\mathcal B_t$, DOCO first computes a candidate transition from the current state $s_t$. A history operator $\mathcal H$ adjusts its compensation increment using the preceding executed increment $\mathbf v_{t-1}$ and an accumulated direction $\mathbf m_{t-1}$:
\begin{equation}
\begin{aligned}
\widetilde s_{t+1}&=\mathcal U_{\mathrm{DOCO}}(s_t,\mathcal B_t),\\
s_{t+1}&=\mathcal H(s_t,\widetilde s_{t+1};
\mathbf v_{t-1},\mathbf m_{t-1}).
\end{aligned}
\label{eq:online_update}
\end{equation}
The operator dampens history-opposed components and blends the adjusted increment with the preceding update. It modifies stored prompt and compensation-statistic coordinates; the remaining DOCO state fields follow the candidate transition. Projection, blending and initialisation rules are specified in the supplement.

For the current batch, we compute the candidate from a snapshot of $s_t$, restore that snapshot, and obtain the block-3 features and restored logits before committing $s_{t+1}$ once. Restoration outputs enter neither the adaptation objective nor the subsequent compensation update.

Complete-system comparisons evaluate restoration together with history-stabilised compensation against native DOCO, which retains its own trajectory and readout timing.

Clean-reference margins target known-class discrimination; detection uses the resulting logits.
Denote the restored logits for an input $x$ by $\ell^{\mathrm{rest}}(x)$, with $C=|\mathcal Y^S|$ known classes. We predict $\hat y=\arg\max_{c\in\mathcal Y^S}\ell_c^{\mathrm{rest}}(x)$ and use
\begin{equation}
E(x)=\log\sum_{c=1}^{C}\exp\bigl(\ell_c^{\mathrm{rest}}(x)\bigr)
\label{eq:energy}
\end{equation}
as an ID-oriented detection score: larger values indicate greater evidence for known classes. We evaluate known-class recognition and energy-based unknown detection separately using the restored logits. Predictions require no target labels, clean test references or corruption identities.

\section{Experiments}
\label{sec:experiments}

\subsection{Experimental setup}
\label{sec:exp-setup}
\paragraph{Data and protocols.}
We evaluate known-class recognition on ImageNet-C~\cite{hendrycks2019corruptions} and unknown detection against Places365-C~\cite{zhou2018places}, Textures-C~\cite{cimpoi2014dtd}, iNaturalist-C~\cite{vanhorn2018inaturalist}, SUN-C~\cite{xiao2010sun}, SSB-Hard-C~\cite{vaze2022ssb}, and NINCO-C~\cite{bitterwolf2023ninco}. The source training, source validation, and ImageNet-C evaluation sets contain mutually disjoint image identities. We construct Semantic--Degradation Coupled OCTTA (SDC-OCTTA) to test whether class-discriminative recovery remains useful as semantic-group--corruption associations change under balanced corruption frequencies. It mixes Gaussian noise, motion blur, fog and contrast across four phases, with each corruption contributing one quarter of both ID and OOD inputs per batch. Conventional OCTTA-4C presents the same corruptions sequentially without resets; OCTTA-15C follows DOCO's standard benchmark~\cite{yang2026doco}. We use severity 5 and equal ID/OOD proportions. The supplement specifies the SDC schedule and sampling details.

\paragraph{Baselines and comparisons.}
We compare with Source, Tent~\cite{wang2021tent}, CoTTA~\cite{wang2022cotta}, EATA~\cite{niu2022eata}, SAR~\cite{niu2023sar}, OSTTA~\cite{lee2023ostta}, S-COME, E-COME~\cite{zhang2025come}, STAMP~\cite{yu2024stamp}, UniEnt~\cite{gao2024unient}, DPCore~\cite{zhang2025dpcore} and native DOCO~\cite{yang2026doco}. We use three comparisons: the \emph{native DOCO diagnostic} tracks its objective and class margins across its own updates; \emph{matched-state restoration} compares restored and unmodified readouts at identical history-stabilised states; and \emph{complete-system evaluation} compares full \method{}, including history-stabilised compensation and frozen restoration, against native DOCO and other baselines.

\paragraph{Metrics and implementation.}
We report ID top-1 accuracy (ACC), ID-positive detection AUROC (AUC) and H-score:
\begin{equation}
H=\frac{2AU}{A+U},
\label{eq:hscore}
\end{equation}
Here $A$ and $U$ are ACC and AUROC. Following DOCO, H-scores are calculated before macro-averaging across six OOD evaluations. We use an ImageNet-pretrained ViT-B/16~\cite{dosovitskiy2021vit}. Source fitting, calibration and online implementation details are given in the supplement.

\subsection{Class discrimination after domain compensation}
\label{sec:mechanism-gap}
During native DOCO adaptation, lower compensation loss can coexist with worse class margins. On a conventional OCTTA-4C diagnostic stream, its full objective and statistical alignment term decrease at all 27 audited update checkpoints, while the structural regulariser increases. With checkpoint-matched batches and fixed inferred ID support, true-class margins decrease in 51.9\% of 865 paired ID diagnostic records, and clean-referenced relation errors increase in 53.6\%.

\paragraph{Class errors after correct ID routing.}
Under history-stabilised compensation without restoration in the SDC stream, 4,239 of 16,668 true ID prediction records correctly routed as ID remain misclassified (25.4\%). We examine how spatial correction changes these class assignments.

\subsection{Recovering residual class errors}
\label{sec:restoration-recovery}
\paragraph{Matched-state prediction changes.}
Under matched history-stabilised states before the next update is committed, enabling the frozen restorer improves ID accuracy, AUC and H-score across all six SDC-OCTTA OOD evaluations, with mean gains of 5.5, 2.2 and 4.4 percentage points, respectively (\cref{tab:matched-spatial-restoration}).

\begin{table}[!htbp]
\centering\small
\setlength{\tabcolsep}{7pt}
\caption{\textbf{Per-OOD effect of spatial restoration on SDC-OCTTA} (percentage points). Each entry reports full \method{} minus w/o Restoration under identical history-stabilised compensation states.}
\label{tab:matched-spatial-restoration}
\begin{tabular*}{\columnwidth}
{@{\extracolsep{\fill}}lrrr@{}}
\toprule
OOD dataset & $\Delta\mathrm{ACC}$ & $\Delta\mathrm{AUC}$ & $\Delta H$\\
\midrule
Places365-C & +5.3 & +0.6 & +3.4\\
Textures-C & +6.1 & +2.6 & +5.1\\
iNaturalist-C & +5.2 & +2.5 & +4.7\\
SUN-C & +5.2 & +2.0 & +4.3\\
SSB-Hard-C & +5.8 & +3.3 & +4.6\\
NINCO-C & +5.5 & +2.4 & +4.4\\
\midrule
Mean & \textbf{+5.5} & \textbf{+2.2} & \textbf{+4.4}\\
\bottomrule
\end{tabular*}
\end{table}

Among true ID records correctly routed as ID, restoration recovers 368 net correct decisions (\cref{tab:e1-routing}): it corrects 636 of 4,239 initially wrong predictions and changes 268 initially correct predictions to wrong ones. Across all true ID records, including those routed as OOD, it corrects 2,274 predictions and harms 745.

\begin{table}[!htbp]
\centering\small
\setlength{\tabcolsep}{5pt}
\caption{\textbf{Matched-state routing-conditioned ID classification changes} on SDC-OCTTA. Both readouts use identical history-stabilised states. True ID records are grouped by their initial inferred ID/OOD status. Corr. and Harm denote wrong-to-correct and correct-to-wrong class prediction changes; CR and HR are the corresponding rates (\%), using initially wrong and correct predictions as denominators.}
\label{tab:e1-routing}
\begin{tabular*}{\columnwidth}
{@{\extracolsep{\fill}}lrrrrr@{}}
\toprule
Routed as & Wrong & Corr. & Harm & CR$\uparrow$ & HR$\downarrow$\\
\midrule
ID & 4,239 & 636 & 268 & 15.0 & 2.2\\
OOD & 8,085 & 1,638 & 477 & 20.3 & 16.5\\
\midrule
\textbf{All true ID} & 12,324 & 2,274 & 745 & 18.5 & 4.9\\
\bottomrule
\end{tabular*}
\end{table}

\paragraph{Recovery towards clean-reference class relations.}
In a separate conventional OCTTA-4C fixed-state diagnostic, restoration reduces clean-referenced relation error from 2.348 to 1.990 (15.2\% relative reduction), with the largest absolute reduction under motion blur. Clean counterparts provide an offline reference for this diagnostic.

\paragraph{Recovery across changing conditions.}
Across SDC phases $I_1$, $A$, $B$ and $I_2$, matched-state restoration improves ID accuracy by 9.2, 4.7, 4.4 and 3.8 percentage points, respectively. The largest gain occurs in the independent phase $I_1$, and gains continue as class--corruption associations change. Positive net correction occurs in 26 of 32 coupled phase--semantic-group cells. The supplement reports the full group-wise results.

\subsection{What makes restoration effective?}
\label{sec:restoration-ablations}
\paragraph{Clean-reference supervision.}
On SUN-C SDC-OCTTA, we compare no restoration, CE-only restoration and CE+CleanRelation at matched history-stabilised states (\cref{tab:cleanrelation}). Training the spatial restorer with classification supervision already improves ID accuracy from 56.2\% to 60.4\%. Adding CleanRelation raises it to 61.4\% and increases H-score from 70.4\% to 71.2\%. This ablation supports their complementary roles: cross-entropy learns useful corrections, and matching selected clean-reference margins improves them further.

\begin{table}[!htbp]
\centering\small
\setlength{\tabcolsep}{3pt}
\caption{\textbf{Training supervision on SUN-C SDC-OCTTA} (\%). No restoration, CE-only restoration and CE+CleanRelation are compared at matched history-stabilised online states.}
\label{tab:cleanrelation}
\begin{tabular*}{\columnwidth}{@{\extracolsep{\fill}}lccrrr@{}}
\toprule
Variant & Restorer & CleanRel & ACC$\uparrow$ & AUC$\uparrow$ & H$\uparrow$\\
\midrule
w/o Restoration & $\times$ & -- & 56.2 & 82.7 & 66.9\\
CE-only & $\checkmark$ & $\times$ & 60.4 & 84.3 & 70.4\\
\textbf{Full \method{}} & $\checkmark$ & $\checkmark$ &
\textbf{61.4} & \textbf{84.7} & \textbf{71.2}\\
\bottomrule
\end{tabular*}

\end{table}

\paragraph{Correction direction and spatial organisation.}
On the conventional OCTTA-4C diagnostic stream, we hold online states, strength and per-input residual norm fixed, then compare the learned residual with its reversal, a norm-matched random perturbation and patch shuffling (\cref{tab:restoration-controls}). The learned correction gives the lowest relation error and highest ID accuracy. Reversal sharply degrades recognition, and shuffling reduces the gain. At equal residual magnitude, the benefit therefore depends on correction direction and spatial organisation, consistent with the class-margin sensitivity in \cref{sec:class_margin}. The supplement reports the strength sweep.

\begin{table}[!htbp]
\centering\small
\setlength{\tabcolsep}{3pt}
\caption{\textbf{Restoration-direction controls} at fixed online states and $\alpha=0.5$ on the four-corruption diagnostic stream. Each control preserves the per-input residual norm. CR/HR are wrong-to-correct/correct-to-wrong rates (\%); Rel. err. is clean-referenced relation error.}
\label{tab:restoration-controls}
\begin{tabular*}{\columnwidth}{@{\extracolsep{\fill}}lrrrr@{}}
\toprule
Direction & Rel. err.$\downarrow$ & CR$\uparrow$ & HR$\downarrow$ & ACC$\uparrow$\\
\midrule
\textbf{Learned $d$} & \textbf{2.0} & \textbf{15.7} & 5.1 & \textbf{65.0}\\
Reversed $-d$ & 3.0 & 5.8 & 21.6 & 51.0\\
Norm-matched random & 2.4 & 4.5 & \textbf{4.6} & 61.1\\
Patch-shuffled $d$ & 2.2 & 11.6 & 5.4 & 63.3\\
\bottomrule
\end{tabular*}
\end{table}

\begin{figure*}[!t]
    \centering
    \includegraphics[width=\textwidth]{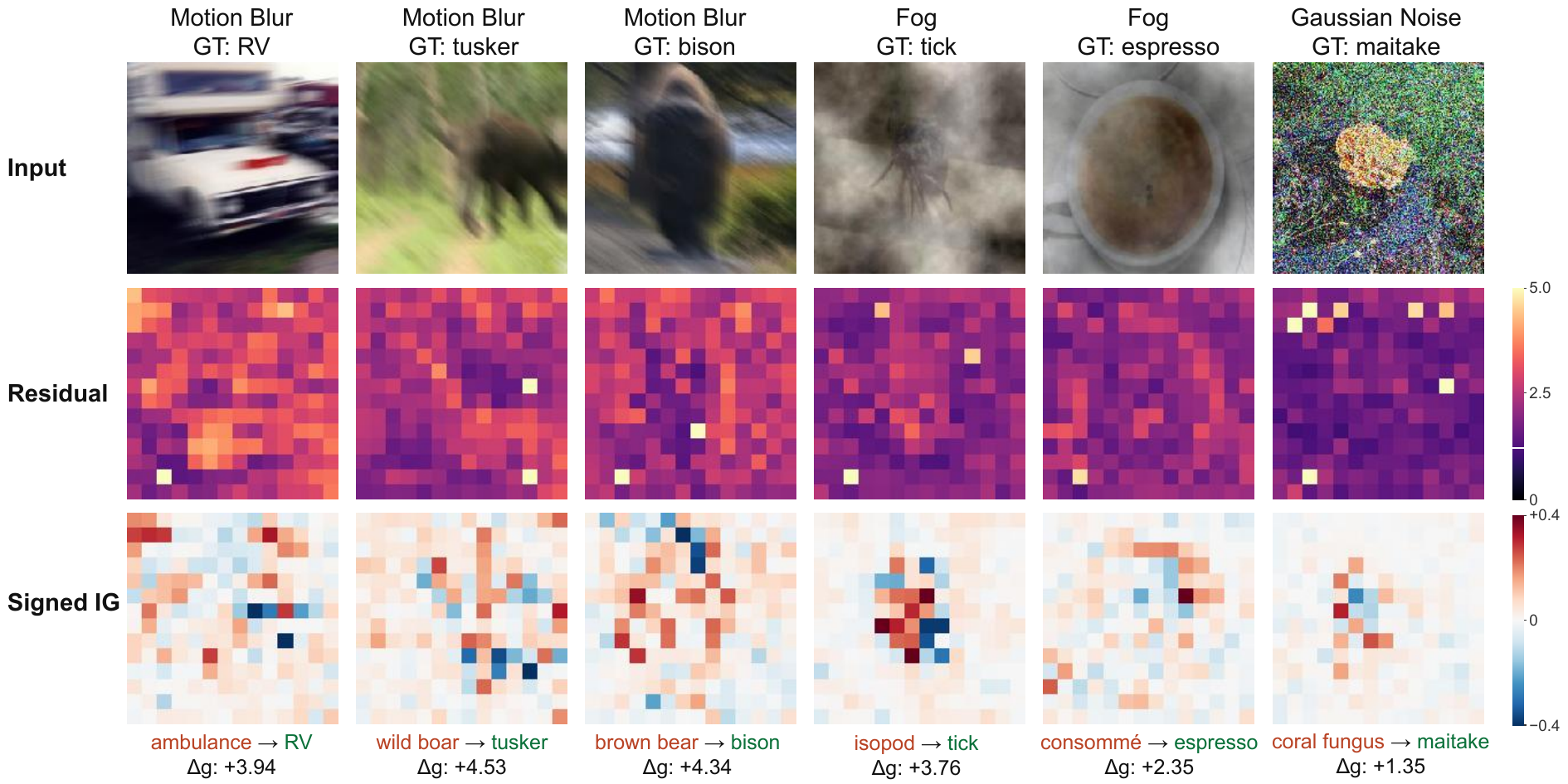}
    \caption{\textbf{Spatially heterogeneous contributions of restoration to class margins.} Six selected wrong-to-correct ImageNet-C ID examples from the SUN-C SDC-OCTTA evaluation. Rows show corrupted inputs, magnitudes of the residual injected into ViT Block-3 patch tokens, and signed integrated-gradient (IG) contributions to the fixed-rival logit-gap change along the restoration OFF-to-ON path. Red and blue indicate positive and negative IG contributions. For each example, both readouts use the same pre-update online state. Bottom labels show prediction changes and $\Delta g=g_{\mathrm{ON}}-g_{\mathrm{OFF}}$, where $g=\ell_y-\ell_r$ with rival $r$ fixed across both readouts. Each heatmap type uses a shared color scale across examples.}
    \label{fig:spatial-evidence}
\end{figure*}

\paragraph{Spatial evidence of restoration.}
To investigate how correction magnitude relates to class-discriminative contributions, we examine six selected wrong-to-correct ID examples from the SUN-C SDC-OCTTA evaluation (\cref{fig:spatial-evidence}). We compare restoration OFF and ON at identical pre-update states and use integrated gradients to attribute the resulting true-class-versus-fixed-rival logit-gap change to individual patch corrections.

The residual-magnitude and signed-attribution maps exhibit distinct spatial patterns: large patch-level corrections can contribute little or negatively to the class margin. Together with the norm-matched direction and patch-shuffling controls in \cref{tab:restoration-controls}, these results highlight the importance of correction direction and spatial organisation beyond residual magnitude.

\subsection{Open-set recognition under changing conditions}
\label{sec:exp-sdc}

\begin{figure*}[!t]
    \centering
    \includegraphics[width=\textwidth]{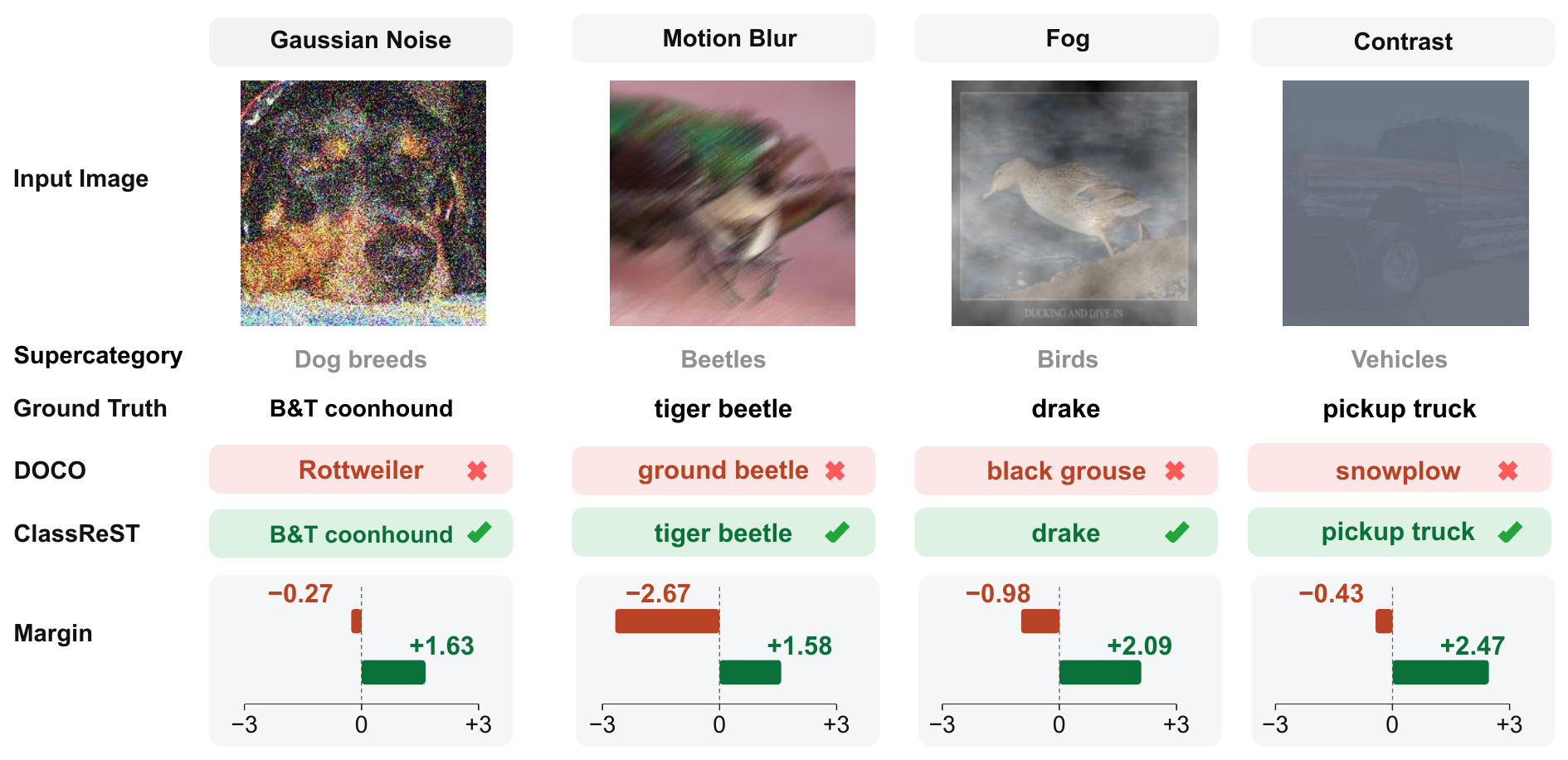}
    \caption{\textbf{Class decisions in the complete-system comparison.}
        Four selected ID inputs from the coupled phases of SDC-OCTTA are shown, with SUN-C as the OOD dataset. Predictions compare native DOCO with full ClassReST, including history-stabilised compensation. Each margin is the ground-truth logit minus the strongest competing logit, computed separately for each system. B\&T denotes black-and-tan.}
    \label{fig:qualitative}
\end{figure*}

\paragraph{Complete-system SDC-OCTTA results.}
Across six OOD datasets, full \method{} achieves 61.0\% ACC, 78.1\% AUC and 68.1\% H-score, improving on DOCO by 6.9, 5.1 and 6.2 percentage points, respectively (\cref{tab:sdc-main}). H-score improves for every evaluated OOD dataset. Full per-OOD and worst-condition comparisons are provided in the supplement.

\begin{table}[!htbp]
\centering\small
\setlength{\tabcolsep}{5pt}
\caption{\textbf{Complete-system SDC-OCTTA averages} across six OOD evaluations (\%). Alongside Source, native DOCO and full \method{}, we show the other baselines with the highest mean ACC/H-score (EATA) and AUC (DPCore). The full per-OOD comparison is in the supplement.}
\label{tab:sdc-main}
\begin{tabular*}{\columnwidth}{@{\extracolsep{\fill}}lrrr@{}}
\toprule
Method & ACC$\uparrow$ & AUC$\uparrow$ & H$\uparrow$\\
\midrule
Source & 38.5 & 57.2 & 46.0\\
EATA & 53.5 & 64.8 & 58.5\\
DPCore & 48.4 & 71.9 & 57.4\\
Native DOCO & 54.1 & 73.0 & 61.9\\
\midrule
\textbf{Full \method{}} & \textbf{61.0} & \textbf{78.1} & \textbf{68.1}\\
\bottomrule
\end{tabular*}
\end{table}

\cref{fig:qualitative} shows four selected ID inputs that native DOCO misclassifies and full \method{} classifies correctly with positive correct-class margins.

\paragraph{Conventional corruption sequences.}
Under conventional OCTTA-4C, full \method{} attains 64.7\% ACC, 82.0\% AUC and 72.1\% H-score, compared with 61.9\%, 80.9\% and 69.8\% for native DOCO. On OCTTA-15C, the complete system achieves a mean H-score of 72.6\%, compared with DOCO's published reference of 70.1\%; AUC increases on four OOD datasets and decreases slightly on SUN-C and SSB-Hard-C. The supplement gives both per-OOD comparisons.

\FloatBarrier
\section{Conclusion}
\label{sec:conclusion}

Matching feature statistics and preserving relationships among examples do not ensure correct class decisions. We use clean-reference source supervision to guide input-dependent spatial restoration without target-time labels. Controlled experiments isolate residual class-error correction under identical compensation states, while complete-system evaluation shows gains in known-class recognition and unknown detection under the tested corruption conditions.

Our findings suggest a broader design principle for visual recognition under distribution shift: use online domain compensation to respond to changing visual conditions, and source-learned class relationships to guide the recovery of discriminative evidence. The two mechanisms address different aspects of reliable recognition and can operate together during deployment.
\section*{Acknowledgments}
This work was supported by the Top-Ten Technological Projects for Hunan Province (Grant No. 2026QK1007). We acknowledge the computational resources provided by the High Performance Computing Center of Central South University.

\section*{LLM Usage Statement}
We used OpenAI ChatGPT for reviewer-style feedback and to assist with manuscript editing, related-work discussions, mathematical exposition, experimental design and analysis, figure refinement, dataset preparation, and research discussions. OpenAI Codex was used to generate portions of the experimental code, which the authors reviewed and refined. The authors take full responsibility for the research decisions, scientific claims, experimental protocols, and reported results.

{\small\bibliographystyle{ieeenat_fullname}\bibliography{main}}
% Submit supplementary as a separate PDF. It is NOT included in this main PDF.
\end{document}